\documentclass[11pt]{article}
\usepackage[ansinew]{inputenc}
\usepackage{amsmath,amssymb}
\usepackage{color}
\usepackage{a4wide}
\usepackage{verbatim}
\usepackage{scrtime}
\usepackage{float}
\usepackage{graphicx}
\usepackage{graphics}
\usepackage{pstricks}
\usepackage{fancyhdr}
\usepackage{subfigure}
\usepackage{amsthm}
\usepackage{float}

\large\normalsize

\newcommand{\trace}{{\rm Tr}}

\newcommand{\rank}{{\rm rank}}
\newcommand{\Exp}{{\mathrm{Exp}}}
\newcommand{\Sym}{{\mathrm{Sym}}}

\newcommand{\subject}{\mathrm{subject\  to}}
\newcommand{\rc}{\nabla}
\newcommand{\D}{\mathrm{D}}

\newcommand{\GL}[1]{{\mathrm{GL}({#1})}}

\newcommand{\set}[2]{{\{{#1}:\ {#2}\}}}

\newcommand{\grad}{\mathrm{grad}}
\newcommand{\hess}{\mathrm{Hess}}

\newcommand{\mat}[1]{{\bf #1}}

\newcommand*\samethanks[1][\value{footnote}]{\footnotemark[#1]}

\renewcommand{\subject}{\mathrm{subject\  to}}
\renewcommand{\min}{\mathrm{min}}

\title{A Riemannian geometry for low-rank matrix completion\thanks{This paper presents research results of the Belgian Network DYSCO (Dynamical Systems, Control, and Optimization), funded by the Interuniversity Attraction Poles Programme, initiated by the Belgian State, Science Policy Office. The scientific responsibility rests with its authors. Bamdev Mishra is a research fellow of the Belgian National Fund for Scientific Research (FNRS).}
}
\author{B.~Mishra\thanks{Department of Electrical Engineering and Computer Science, University of Li\`ege, 4000 Li\`ege,
Belgium (B.Mishra@ulg.ac.be, R.Sepulchre@ulg.ac.be).}
        \and K.~Adithya Apuroop \thanks{Department of Electrical Engineering, Indian Institute of Technology Bombay, Powai, Mumbai, Postcode 400076, Maharashtra, India (adithya.apuroop@iitb.ac.in).}
        \and R.~Sepulchre\samethanks[2]
        }

\begin{document}

\maketitle

\begin{abstract}
We propose a new Riemannian geometry for fixed-rank matrices that is specifically tailored to the low-rank matrix completion problem. Exploiting the degree of freedom of a quotient space, we tune the metric on our search space to the particular least square cost function. At one level, it illustrates in a novel way how to exploit the versatile framework of optimization on quotient manifold. At another level, our algorithms can be considered as improved versions of LMaFit, the state-of-the-art Gauss-Seidel algorithm. We develop necessary tools needed to perform both first-order and second-order optimization. In particular, we propose gradient descent schemes (steepest descent and conjugate gradient) and trust-region algorithms. We also show that, thanks to the simplicity of the cost function, it is numerically cheap to perform an exact linesearch given a search direction, which makes our algorithms competitive with the state-of-the-art on standard low-rank matrix completion instances.

\end{abstract}

\section{Introduction}
The problem of low-rank matrix completion amounts to completing a matrix from a small number of entries by assuming a low-rank model for the matrix. This problem has been addressed both from the theoretical \cite{candes08b, gross11a} as well as algorithmic viewpoints \cite{rennie05a, cai08a, lee09a, meka09a, keshavan10a, simonsson10a, jain10a, mazumder10a, boumal11a, meyer11a, mishra11a, mishra12a}. A standard way of approaching the problem is by casting the low-rank matrix completion problem as a fixed-rank optimization problem with the assumption that the optimal rank $r$ is known a priori as shown below
\begin{equation}\label{eq:matrix-completion-formulation}
\begin{array}{llll}
	\min_{\mat{X}\in\mathbb{R}^{n \times m}}
		&	\frac{ 1}{|\Omega|}\|\mathcal{P}_{\Omega}(\mat{X}) - \mathcal{P}_{\Omega}(\widetilde{\mat{X}})\|_F^2 \\
	 \text{subject to} & \rank(\mat{X})=r,
\end{array}
\end{equation}
where $\widetilde{\mat{X}}\in\mathbb{R}^{n\times m}$ is a matrix whose entries known for indices if they belong to the subset $(i,j)\in\Omega$, where $\Omega$ is a subset of the complete set of indices $\{(i,j):i\in\{1,...,n\}\text{ and }j\in\{1,...,m\}\}$. The operator $\mathcal{P}_{\Omega}(\mat{X}_{ij})=\mat{X}_{ij}$ if $(i,j) \in \Omega$ and $\mathcal{P}_{\Omega}(\mat{X}_{ij})=0$ otherwise is called the \emph{orthogonal sampling operator} and is a mathematically convenient way to represent the subset of entries. The objective function is, therefore, a mean least square objective function where $\|\cdot \|_F$ is \emph{Frobenius} norm with $|\Omega|$ is the cardinality of the set $\Omega$ (equal to the number of known entries). The search space $\mathbb{R}_r^{n \times m}$ is the space of $r-\rank$ matrices of size $n \times m$ , with $r \ll \min\{ m, n \}$.

The rank constraint correlates the known with the unknown entries. The number of given entries $|\Omega|$ is typically much smaller than $n m$, the total number of entries in $\widetilde{\mat{X}}$. Recent contributions provide conditions on $|\Omega|$ under which exact reconstruction is possible from entries sampled uniformly and at random \cite{candes08b,cai08a,keshavan10a}. 

A popular way to tackle the rank-constraint in (\ref{eq:matrix-completion-formulation}) is by using a factorization model. In this paper, we pursue our research on the geometry of factorization models. It builds upon earlier works \cite{meyer11b, meyer11a, mishra12a} which describe factorization models and optimization on the resulting search space. The geometries described in \cite{meyer11b, mishra12a} only deal with quotient structure that results from the symmetries of factorization models. Here we move a step forward by tuning the metric to the cost function. The resulting geometry reduces to the state-of-the-art algorithm LMaFit \cite{wen10a} as a special case of our optimization scheme. We develop tools necessary to propose efficient first-order and second-order optimization algorithms with the new geometry. Our simulations suggest that the proposed algorithms scale to large dimensions and compete favorably with the state-of-the-art.

At the time of submission of this paper, the authors were alerted (by Google Scholar) of the preprint \cite{ngo12a}. At the conceptual level, the two papers present the same idea, each with its own twist. The twist of the present contribution is to emphasize that the new metric is one particular selection among Riemannian quotient geometries of low-rank matrices. It underlines the value of a geometric framework that parameterizes the degrees of freedom of the metric (the optimization of which is problem dependent) while providing a general theory of convergence and algorithms (the collection of which is problem independent).

\section{Exploiting the problem structure}
Efficient algorithms depend on properly exploiting both the structure of the constraints and the structure of the cost function. A natural way to exploit the rank constraint is through matrix factorization, which leads to a quotient structure of the search space \cite{meyer11b, mishra12a}. We review one of the geometries below. The structure of the cost function is then exploited to select one particular metric on the quotient manifold.

\subsection*{The quotient nature of matrix factorization}
 A $r-{\rm rank}$ matrix $\mat{X} \in \mathbb{R}^{n \times m}_r$ is factorized as 
\begin{equation}\label{eq:factorization_gh}
\mat{X} = \mat{G} \mat{H}^T
\end{equation}
where $\mat{G} \in \mathbb{R}_* ^{n \times r}$ and $\mat{H} \in \mathbb{R}^{m \times r}_*$ are full column-rank matrices. Such a factorization is not unique as $\mat{X}$ remains unchanged by \emph{scaling} the factors
\begin{equation}\label{eq:symmetry_gh}
(\mat{G},\mat{H})\mapsto (\mat{G}\mat{M}^{-1},\mat{H}\mat{M}^{T}),
\end{equation}
for any non-singular matrix $\mat{M} \in \GL{r}$, the set of $r \times r$ non-singular matrices and we have $\mat{X} = \mat{G} \mat{H}^T = \mat{G} \mat{M}^{-1}  (\mat{H}   \mat{M}^T)^T$ \cite{piziak99a}. 

The classical remedy to remove this indeterminacy is Cholesky factorization, which requires further (triangular-like) structure in the factors. LU decomposition is a way forward \cite{golub06a}. In contrast, we encode the invariance map \eqref{eq:symmetry_gh} in an abstract search space by optimizing over a set of equivalence classes defined as 
\begin{equation}\label{eq:equivalence-classes-balanced}
 [(\mat{G},\mat{H})] = \set{(\mat{G}\mat{M}^{-1},\mat{H}\mat{M}^{T})}{\mat{M}\in\mathrm{GL}(r)}.
\end{equation}
The set of equivalence classes is termed as the quotient space of $\overline{\mathcal{M}}_r$ by $\GL{r}$ and is denoted as 
\begin{equation}\label{eq:quotient-balanced}
	{\mathcal{M}}_r:= \overline{\mathcal{M}}_r /\GL{r},
\end{equation}
where the total space $\overline{\mathcal{M}}_r$ is the product space $\mathbb{R}_*^{n \times r} \times \mathbb{R}_*^{m \times r}$.

To do optimization on the abstract search space $\mathcal{M}_r$ we need to select a metric on the total space $\overline{\mathcal{M}}_r$ such that the quotient search space is a Riemannian submersion, Section $6.3.2$ of \cite{absil08a}. The metric should make the inner product between tangent vectors invariant along the equivalence class $ [(\mat{G},\mat{H})]$. This is the only constraint imposed by the geometry of the search space.	

\subsection*{Scaling}
A further selection constraint for the metric is to look at the Hessian of the objective function. The Newton algorithm is a \emph{scaled} gradient descent algorithm, in which the search space is endowed with a Riemannian metric induced by the Hessian of the cost function \cite{nesterov03a, nocedal06a}. In many cases, using the full Hessian information is computationally costly. A popular compromise between convergence and numerical efficiency is to scale the gradient by the diagonal elements of the Hessian \cite{nocedal06a}. If we vectorize (column-on-column) $\mat{G} \rightarrow g$ and $\mat{H} \rightarrow h$ and stack them on another like 
\[
\left [
\begin{array}{llll} 
g \\
h	
\end{array}
\right ]
 \]
then the full Hessian will be a symmetric $(n+m)r \times (n+m)r$ matrix but not necessarily positive definite as the objective function is non-convex in the variables $g$ and $h$ together. However, the diagonal elements are going to be strictly positive because, fixing $h$, the objective function is strictly convex in $g$ and \emph{vice versa}. 

The full Hessian for the matrix completion function is given in \cite{buchanan05a} where the authors deal with non-positive definiteness of the Hessian by shifting the eigenvalues using the Levenberg-Marquardt optimization method \cite{nocedal06a}. However, the resulting optimization algorithm is numerically costly and unsuitable for large dimensions \cite{chen08a}. Hence, we avoid using the full Hessian information and instead concentrate on the diagonal elements of the Hessian of the \emph{matrix approximation} function 
\begin{equation}\label{eq:matrix_approximation}
 \|  \widetilde{\mat{X}} - \mat{GH}^T  \|_F  ^2
\end{equation}
Minimizing this function with respect to $\mat{G}$ and $\mat{H}$ amounts to learning the dominant $r-{\rm rank}$ subspace of $\widetilde{\mat{X}}$ \cite{golub06a}. The diagonal elements of the Hessian of the matrix approximation cost function  (which number $(n + m)r$) have a simple form. Let 
\[
\left [
\begin{array}{lll}
{\rm Hess}_g & \mat{0} \\
\mat{0} & {\rm Hess}_h
\end{array}
\right ]
\]
be the diagonal approximation of the full Hessian of the matrix approximation function where ${\rm Hess}_g$ is a diagonal matrix size $nr \times nr$ and ${\rm Hess}_h$ is a diagonal matrix of size $mr\times mr$. Scaling the gradients (with respect to vectorized formulation) with the diagonal elements of the Hessian \cite{nocedal06a} would then mean following transformation 
\[
\begin{array}{llll}
\frac{\partial}{\partial g}  \longrightarrow {\rm Hess}_g^{-1}  \  \frac{\partial}{\partial g}  \\ 
\frac{\partial}{\partial h}   \longrightarrow {\rm Hess}_h^{-1}  \  \frac{\partial}{\partial h}  \\
\end{array}
\]
where $\frac{\partial}{\partial g} \in \mathbb{R}^{nr}$ and $\frac{\partial}{\partial h} \in \mathbb{R}^{mr} $ are the Euclidean partial derivatives in vectorized form. In matrix form (just rearranging the terms) and substituting ${\rm Hess}_g$ and ${\rm Hess}_h$ appropriately, we have the following equivalent transformation 
\begin{equation}\label{eq:diagonal_scaling}
\begin{array}{llll}
\frac{\partial}{\partial \mat{G}}  \longrightarrow  ( \frac{\partial}{\partial \mat{G}} ) [ {\rm diag}(\mat{H}^T\mat{H}) ]^{-1} \\
\frac{\partial}{\partial \mat{H}}   \longrightarrow  (\frac{\partial}{\partial \mat{H}} )  [ {\rm diag}(\mat{G}^T\mat{G}) ]^{-1}\\
\end{array}
\end{equation}
where $\frac{\partial}{\partial \mat{G}} $ and $\frac{\partial}{\partial \mat{H}} $ are the Euclidean partial derivatives with respect to matrices $\mat{G}$ and $\mat{H}$.  ${\rm diag(\mat{H}^T \mat{H})}$ is a diagonal matrix extracting the diagonal of $\mat{H}^T \mat{H}$. This seems to be a suitable scaling for a gradient descent algorithm for minimizing the matrix approximation problem (\ref{eq:matrix_approximation}). 

This also seems to be a reasonable scaling for algorithms directed towards the matrix completion problem (\ref{eq:matrix-completion-formulation}). This is because, when we complete a low-rank matrix, we implicitly try to learn the dominant subspace, by learning on a smaller set of known entries. 

\subsection*{A new metric on the tangent space}
The tangent space of the total (product) space $\overline{\mathcal M}_r$ has the following characterizations due to the product structure $T_{\bar{x}} \overline{\mathcal{M}}_r$ at $\bar{x} = (\mat{G}, \mat{H}) \in \overline{\mathcal M}_r$ has the expression
\[
T_{\bar{x}} \overline{\mathcal{M}}_r = \mathbb{R}^{n \times r} \times \mathbb{R}^{m \times r}.
\]
To choose a \emph{proper} metric on the total space $\overline{\mathcal{M}}_r$ we need to take into account the \emph{symmetry} (\ref{eq:symmetry_gh}) imposed by the factorization model and \emph{scaling}, similar to (\ref{eq:diagonal_scaling}). The new metric, we propose is 
\begin{equation}\label{eq:metric_gh}
\begin{array}{lll}
\bar{g}_{\bar{x}}
 (    \bar{  \xi}_{\bar{x}} ,  \bar{\eta}_{\bar{x}}  )  & =  &\trace  ((\mat{H}^T\mat{H})\bar{\xi}_{\mat G}^T  \bar{\eta}_{\mat G})   +  \trace ( (\mat{G}^T\mat{G}) \bar{\xi}^T_{\mat H} \bar{\eta}_{\mat H}  ) \\
\end{array}
\end{equation}
where $\bar{x} = (\mat{G}, \mat{H})$ and $\bar{\xi}_{\bar{x}},\bar{\eta}_{\bar{x}} \in T_{\bar{x}} \overline{\mathcal{M}}_r$. Note that this is not the \emph{right-invariant metric} introduced in \cite{meyer11a} which is
\begin{equation}\label{eq:right_inv_metric_gh}
\begin{array}{lll}
\bar{g}_{\bar{x}}
 (    \bar{  \xi}_{\bar{x}} ,  \bar{\eta}_{\bar{x}}  )  & =  &\trace  ((\mat{G}^T\mat{G})^{-1}\bar{\xi}_{\mat G}^T  \bar{\eta}_{\mat G})   +  \trace ( (\mat{H}^T\mat{H})^{-1} \bar{\xi}^T_{\mat H} \bar{\eta}_{\mat H} ). \\
\end{array}
\end{equation}
Both these metrics lead to complete metric spaces and both satisfy the symmetry condition (\ref{eq:symmetry_gh}) but they differ in the scaling. The right-invariant metric is motivated purely by the geometry of the product space. It is not tuned to a particular cost function. On the other hand, our choice is motivated by the geometry as well as the particular least square cost function of the matrix completion problem. With the right-invariant metric and our new metric, the scaling of the gradients are characterized as 
\begin{equation}\label{eq:difference_metric}
\begin{array}{llll}
{\rm New\ metric:}& \frac{\partial}{\partial \mat{G}}  \longrightarrow  ( \frac{\partial}{\partial \mat{G}} )  (\mat{H}^T\mat{H}) ^{-1}, & \frac{\partial}{\partial \mat{H}}   \longrightarrow  (\frac{\partial}{\partial \mat{H}} ) (\mat{G}^T\mat{G}) ^{-1}\\
{\rm Right-invariant\ metric:}& \frac{\partial}{\partial \mat{G}}  \longrightarrow  ( \frac{\partial}{\partial \mat{G}} )  (\mat{G}^T\mat{G}), & \frac{\partial}{\partial \mat{H}}   \longrightarrow  (\frac{\partial}{\partial \mat{H}} ) (\mat{H}^T\mat{H})\\
\end{array}
\end{equation}
The scaling obtained by our metric in some way inverse to that by the right-invariant metric. Consider the case when $\mat{G}^T\mat{G} = \mat{H}^T\mat{H}$ (balancing update as called in \cite{meyer11b}), the scaling is indeed inverse. In fact comparing (\ref{eq:metric_gh}) and (\ref{eq:right_inv_metric_gh}) suggests that the right-invariant metric (\ref{eq:right_inv_metric_gh}) is not well suited to the matrix completion, explaining the poor performance often observed in simulation (in particular instances as shown later in Section \ref{sec:comparison_gd} ). The proposed new metric (\ref{eq:metric_gh}) tries to take into account the scaling, as much as possible.

\section{Optimization related ingredients}
Once the metric (\ref{eq:metric_gh}) is chosen on the total space $\overline{\mathcal M}_r$, the discussion relating the quotient space ${\mathcal M}_r$ with the new metric follows directly from \cite{absil08a}. To perform optimization on the search space we need the notions of the following concepts. These can be directly derived by following the steps in \cite{absil08a}.

\subsection*{Notions of optimization on the quotient space}
For quotient manifolds $\mathcal{M}_r=\overline{\mathcal{M}}_r/\sim$ a tangent vector $\xi_{x} \in T_{x}\mathcal{M}_r$ at $x=[\bar{x}]$ is restricted to the directions that do not induce a displacement along the set of equivalence classes. This is achieved by decomposing the tangent space in the total space $T_{\bar{x}}\overline{\mathcal{M}}_r$ into complementary spaces
\begin{equation*}
	T_{\bar x}\overline{\mathcal{M}}_r = \mathcal{V}_{\bar x}\overline{\mathcal{M}}_r \oplus   \mathcal{H}_{\bar x}\overline{\mathcal{M}}_r.
\end{equation*}
\begin{figure}[!ht]
 	\centering
 	\begin{tabular}{cc}
 		\includegraphics[scale=.50]{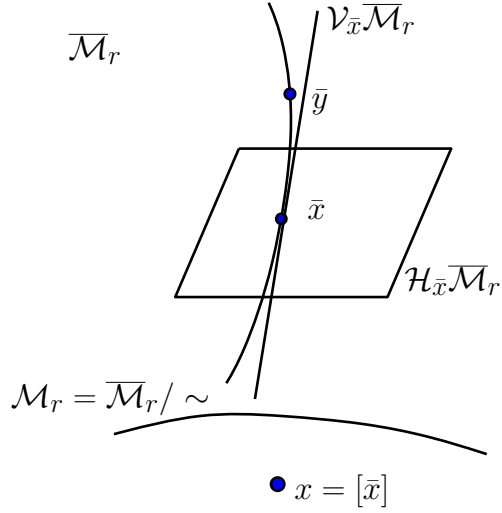}
 	\end{tabular}
 	\caption{The search space is the abstract quotient space $\mathcal{M}_r$. The points $\bar{y}$ and $\bar{x}$ in $\overline{ \mathcal{M}}_r$ belonging to the same equivalence class are represented by a single point $[x]$ in the quotient space $\mathcal{M}_r$.}
 	\label{fig:quotient_optimization}
 \end{figure}
Refer Figure \ref{fig:quotient_optimization} for a graphical illustration. The \emph{vertical space} $\mathcal{V}_{\bar{x}}\overline{\mathcal{M}}_r$ is the set of directions that contains tangent vectors to the equivalence classes. The \emph{horizontal space} $\mathcal{H}_{\bar{x}}\overline{\mathcal{M}}_r$ is the complement of $\mathcal{V}_{\bar{x}}\overline{\mathcal{M}}_r$ in $T_{\bar{x}}\overline{\mathcal{M}}_r$. The horizontal space provides a representation of the abstract tangent vectors to the quotient space, i.e., $T_{x} \mathcal{M}_r$. Indeed, displacements in the vertical space leave the matrix $\mat{X}$ (matrix representation of the point $\bar{x}$) unchanged, which suggests to restrict both tangent vectors and metric to the horizontal space. Once $T_{\bar{x}}\overline{\mathcal{M}}_r$ is endowed with a horizontal distribution $\mathcal{H}_{\bar{x}}\overline{\mathcal{M}}_r$, a given tangent vector $\xi_{x} \in T_{x}\mathcal{M}_r$ at $x$ in the quotient manifold $\mathcal{M}_r$ is uniquely represented by a tangent vector $\bar{\xi}_{\bar{x}}\in\mathcal{H}_{\bar{x}}\overline{\mathcal{M}}_r$ in the total space $\overline{\mathcal{M}}_r$.

The tangent vector $\bar{\xi}_{\bar{x}}\in\mathcal{H}_{\bar{x}}\overline{\mathcal{M}}_r$ is called the \emph{horizontal lift} of $\xi_{x}$ at $\bar{x}$. Provided that the metric defined in the total space is invariant along the set of equivalence classes. A metric $\bar{g}_{\bar{x}}(\bar{\xi}_{\bar{x}},\bar{\zeta}_{\bar{x}})$ in the total space defines a metric $g_x$ on the quotient manifold. Namely,
\begin{equation}\label{eq:metric}
	g_{x}(\xi_{x},\zeta_{x}):=\bar{g}_{\bar{x}}(\bar{\xi}_{\bar{x}},\bar{\zeta}_{\bar{x}})
\end{equation}
where $\xi_x$ and $\zeta_x$ are the tangent vectors in $T_x \mathcal{M}_r$ and $\xi_{\bar{x}}$ and $\zeta_{\bar{x}}$ are their horizontal lifts in $\mathcal{H}_{\bar{x}} \overline{\mathcal{W}}$.

\begin{figure}[!ht]
 	\centering
 	\begin{tabular}{cc}
 		\includegraphics[scale=.50]{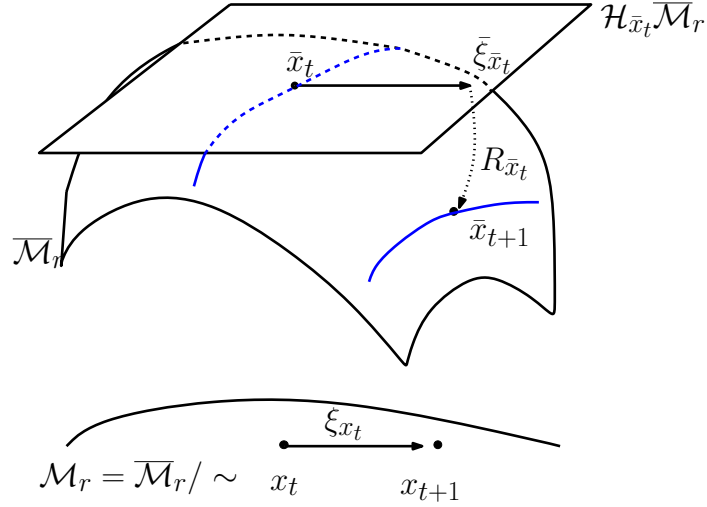}
 	\end{tabular}
 	\caption{A line-search on a Riemannian quotient manifold $\mathcal{M}_r$ in the total space $\overline{\mathcal{M}}_r$. Conceptually, we move on the quotient manifold ${\mathcal M}_r$ from $x_t$ to $x_{t+1}$ but computationally in the total space $\overline{\mathcal{M}}_r$. The search direction is $\bar{\xi}_{\bar{x}_t}$ and belongs to the horizontal space $\mathcal{H}_{\bar{x}_t}\overline{\mathcal{M}}_r$ at point $\bar{x}_{t}$. The retraction mapping maintains feasibility of the iterates. The blue line denotes the equivalence class $[\bar{x}_t]$.}
 	\label{fig:retraction}
 \end{figure}

Natural displacements at $\bar{x}$ in a direction $\bar{\xi}_{\bar{x}} \in \mathcal{H}_{\bar{x}} \overline{\mathcal{W}}$ on the manifold are performed by following geodesics (paths of shortest length on the manifold) starting from $\bar{x}$ and tangent to $\xi_{\bar{x}}$. This is performed by means of the exponential map
\begin{equation*}\label{eq:exponential-map}
	\bar{x}_{t+1} = \Exp_{\bar{x}_t}(s_t \bar{\xi}_{\bar{x}_t}),
\end{equation*}
which induces a line-search algorithm along geodesics with the step size $s_t$. However, the geodesics are generally expensive to compute and, in many cases, are not available in closed-form. A more general update formula is obtained if we relax the constraint of moving along geodesics. The retraction mapping $R_{\bar{x}_t}(s_t\xi_{\bar{x}_t})$ at $\bar{x}_t$ locally approximates the exponential mapping \cite{absil08a}. It provides a numerically attractive alternative to the exponential mapping in the design of optimization algorithms on manifolds, as it reduces the computational complexity of the update while retaining the essential properties that ensure convergence results. A generic abstract line-search algorithm is, thus, based on the update formula
\begin{equation}\label{eq:retraction-map}
	\bar{x}_{t+1} = R_{\bar{x}_t}(s_t \bar{\xi}_{\bar{x}_t})
\end{equation}
where $s_t$ is the step-size. A good step-size is computed using the Armijo rule \cite{nocedal06a}.

Similarly, second-order algorithms on the quotient manifold $\mathcal{W}$ are horizontally lifted and solved in the total space $\overline{\mathcal{W}}$. Additionally, we need to be able to compute the directional derivative of gradient along a search direction. This relationship is captured by \emph{an affine connection} $\rc$ on the manifold. The vector field $\rc _\eta \xi $ implies the \emph{covariant derivative} of vector field $\eta$ with respect to the vector field $\xi$. In the case of $\mathcal{W}$ being a Euclidean space, the affine connection is standard 
\[
\left ( \rc _\xi \eta \right )_x = \lim_{t \rightarrow 0} \frac{\eta_{x + t \xi_x } - \eta_x}{t}.
\]
However, for an arbitrary manifold there exists infinitely many different affine connections except for a specific connection called the \emph{Levi-Civita} or \emph{Riemannian} connection which is always unique. The properties of affine connections and Riemannian connection are in Section $5.2$ and Theorem $5.3.1$ of \cite{absil08a}. The Riemannian connection on the quotient manifold $\mathcal{W}$ is given in terms of the connection in the total space $\overline{\mathcal{W}}$ once the quotient manifold has the structure of a \emph{Riemannian submersion} \cite{absil08a}. 

\subsection*{Horizontal space and projection operator}
A matrix representation of the tangent space of the abstract search space $\mathcal{M}_r$ at the equivalence class $x  = [\bar{x}] \in \mathcal{W}$ relies on the decomposition of $T_{\bar{x}} \overline{\mathcal{M}}_r$ into its \emph{vertical} and \emph{horizontal} subspaces.
The tangent space is decomposed into the vertical and horizontal space which have the following expressions
\begin{equation}\label{eq:horizontal_space_gh}
\begin{array}{lll}
\mathcal{V}_{\bar{x}} \overline{\mathcal{M}}_r =  \left \{  \left (   -\mat{G}\mat{\Lambda} , \mat{H}\mat{\Lambda} ^T   \right ) \quad | \quad \mat{\Lambda} \in \mathbb{R}^{r\times r} \right \}\quad {\rm and} \\
\mathcal{H}_{\bar{x}} \overline{\mathcal{M}}_r = \{   \left ( \bar{ \zeta}_{\mat{G}} , \bar{\zeta}_{\mat{H}}   \right )\quad | \quad   \mat{G}^T \zeta_{\mat{G}} \mat{H}^T \mat{H}  = \mat{G}^T\mat{G} \zeta_{\mat{H}}^T \mat{H}     ,\quad \bar{\zeta}_\mat{G} \in \mathbb{R}^{n \times r}, \bar{\zeta}_\mat{H}   \in \mathbb{R}^{m \times r}  \}
 \end{array}
\end{equation}
where $\bar{\zeta}_{\bar{x}} \in T_{\bar{x}} \overline{\mathcal{M}}_r$ and for any ${ \mat \Lambda} \in \mathbb{R}^{r \times r}$. 

Apart from the characterization of the Horizontal space, we need a mapping
$\Pi_{\bar{x}}: T_{\bar{x}} \overline{\mathcal{M}}_r \mapsto \mathcal{H}_{\bar{x}} \overline{\mathcal{M}}_r$ that maps vectors from the tangent space to the horizontal space \cite{absil07a}. Projecting an element $\bar{\eta}_{\bar{x}} \in T_{\bar{x}} \overline{\mathcal{M}}_r$ onto the horizontal space is accomplished with the operator
\begin{equation}\label{eq:projection_gh}
\Pi_{\bar{x}}(\bar{\eta}_{\bar{x}})=  
\begin{array}{ll}
( \bar{\eta}_{\mat{G}} + \mat{G}\mat{\Lambda} , \bar{\eta}_{\mat{H}} - \mat{H\Lambda}^T )
\end{array}
\end{equation}
where ${\mat \Lambda} \in \mathbb{R}^{r \times r}$ is uniquely obtained by noticing that $\Pi_{\bar{x}}(\bar{\eta}_{\bar{x}})$ belongs to the horizontal space described in (\ref{eq:horizontal_space_gh}) and hence, we have
\begin{equation}\label{eq:Lyapunov_gh}
\begin{array}{llll}
& \mat{ G }^T \left ( \bar{\eta}_{\mat{G}} + \mat{G} {\mat \Lambda} \right )  \mat{H}^T \mat{H} & = &\mat{G}^T \mat{G}   \left (    \bar{\eta}_{\mat{H}} - \mat{H} {\mat \Lambda} ^T\right )^T \mat{H} \\

\Rightarrow &  2 \mat{G}^T\mat{G} \mat{\Lambda}  \mat{H}^T\mat{H} & = & \mat{G}^T\mat{G}\eta_{\mat{H}}^T\mat{H}    - \mat{G}^T\eta_{\mat{G}}  \mat{H}^ T \mat{H} \\

\Rightarrow & \mat{\Lambda} & = &  0.5\left [ \eta_{\mat{H}}^T\mat{H} (\mat{H}^ T \mat{H})^{-1} - (\mat{G}^T\mat{G} )^{-1}\mat{G}^T\eta_{\mat{G}} \right ].
\end{array}
\end{equation}
which has a closed form expression. 

\subsection*{Retraction}
A retraction is a mapping that maps vectors in the horizontal space to points on the search space $\mathcal{M}_r$ and satisfies certain conditions (Definition $4.1.1$ in \cite{absil08a}). We choose a simple retraction \cite{absil08a}
\begin{equation}\label{eq:retraction_gh}
\begin{array}{ll}
R_{\mat{G}} (\bar{\eta}_{\mat G}) = \mat{G} + \bar{\eta}_{\mat G} \\
R_{\mat{H}} (\bar{\eta}_{\mat H}) = \mat{H} + \bar{\eta}_{\mat H} \\
\end{array}
\end{equation}
where $\bar{\eta}_{\bar x} \in \mathcal{H}_{\bar x} \overline{\mathcal M}_r $. 

\subsection*{Vector transport}
A vector transport $\mathcal{T}_{\eta_x} \xi_x$ on a manifold $\mathcal{M}_r$ is a smooth mapping that transports the vector $\xi_x$ at $x$ to a vector in the tangent space at $x + \eta _x$ satisfying certain conditions. Refer Definition $8.1.1$ in \cite{absil08a}. When the total space is an open subset of the Euclidean space which is our case, the vector transport is given in terms of the projection operator (\ref{eq:projection_gh}) and is computed as in Section $8.1.4$ in \cite{absil08a}
\[
\overline{(\mathcal{T}_{\eta_x} \xi_x)}_{\bar{x} + \overline{\eta}_{\bar x} } = \Pi_{\bar{x} + \overline{\eta}_{\bar x}} ({\bar \xi}_{\bar x}).
\]

\subsection*{Exact linesearch}
Because of the simplicity of the cost function and simple retraction used, an exact linesearch is numerically feasible. Given a search direction $\bar{\eta} \in \mathcal{H}_{\bar x} \overline{\mathcal M}_r$ at a point $\bar{x} = (\mat{G}, \mat{H})$, the optimal step-size $t^{*}$ is computed by solving the minimization problem
\begin{equation}\label{eq:exact_linesearch}
t^* = {\rm argmin}_t \quad \| P_{\Omega}(\widetilde{\mat{X}}) - (\mat{G} + t \eta_{\mat G}) (\mat{H} + t \eta_{\mat H})^T\|_F ^2.
\end{equation}
Note that this is a degree $4$ polynomial in $t$. The optima are the roots of the first derivative which is a degree $3$ polynomial. Efficient algorithms exist (including closed-form expressions) for finding the roots of a degree $3$ polynomial and hence, finding the optimal $t^*$ is numerically straight forward.

Note also that for a gradient descent algorithm, computing $t^{*}$ is still numerically more costly than using an approximate step-size procedure like adaptive linesearch in \cite{nocedal06a, mishra12a}. Hence, we do not compute the optimal step-size at each iteration for our gradient descent implementation. We use it only in the very beginning to guess a right step-size order and then use our adaptive step-size procedure described in \cite{mishra12a} for subsequent iterations.

However, an optimal step-size has a significant effect on a conjugate gradient algorithm \cite{nocedal06a}. The extra cost of computing the optimal step-size is compensated by a faster rate of convergence. 

\subsection*{Riemannian connection}
For the trust-region scheme we additionally need the notion of \emph{Riemannian connection} in the total space $\overline{\mathcal{M}}_r$ on the search space $\mathcal{M}_r$. The Riemannian connection generalizes the idea of directional derivative of a vector field on the manifold. In particular, we are interested in the notion on the quotient space. The horizontal lift of the Riemannian connection $\nabla_{\eta_x} \xi_x$, the directional derivative of the vector field $\xi_x \in T_{x}\mathcal{M}_r$ in the direction $\eta_x \in T_{x}\mathcal{M}_r$, is given in terms of the Riemannian connection on the product space $\overline{\mathcal{M}}_r$,
\begin{equation}\label{eq:Riemannian_connection}
\overline{\nabla_{\eta_x} \xi_x} = \Pi_{\bar x} \left( \overline{\nabla}_{\overline{\eta_x}} \overline{\xi_x} \right )
\end{equation}
which is the horizontal projection of the Riemannian connection onto the horizontal space, Proposition $5.3.3$ in \cite{absil08a}. It now remains to find out the Riemannian connection on the total space $\overline{\mathcal{M}}_r$. We find the expression by invoking the \emph{Koszul} formula, Theorem $5.3.1$ in \cite{absil08a}. After a routine calculation, the final expression
\begin{equation}\label{eq:connection_total_space}
\begin{array}{llll}
\overline{\rc}_{\overline{\eta}} \overline{\xi} & = & \D \overline{\xi}[\overline{\eta}] + \left( \mat{A}_{\mat{G}}, \mat{A}_{\mat H} \right),\quad {\rm where} \\
\mat{A}_{\mat G} & = & {\overline \xi}_{\mat G}\Sym( \overline{\eta}_{\mat H} ^T \mat{H} )(\mat{H}^T\mat{H})^{-1}     + \overline{\eta}_{\mat G}\Sym( \overline{\xi}_{\mat H} ^T \mat{H} )(\mat{H}^T\mat{H})^{-1}  - \mat{G}\Sym( \overline{\xi}_{\mat H} ^T \overline{\eta}_{\mat H} ) (\mat{H}^T\mat{H})^{-1} \\

\mat{A}_{\mat H} & = & {\overline \xi}_{\mat H}\Sym( \overline{\eta}_{\mat G} ^T \mat{G} )(\mat{G}^T\mat{G})^{-1}     + \overline{\eta}_{\mat H}\Sym( \overline{\xi}_{\mat G} ^T \mat{G} )(\mat{G}^T\mat{G})^{-1}  - \mat{H}\Sym( \overline{\xi}_{\mat G} ^T \overline{\eta}_{\mat G} ) (\mat{G}^T\mat{G})^{-1} 
\end{array}
\end{equation}
and $ \D \overline{\xi}[\overline{\eta}]$ is the classical Euclidean directional derivative and $\Sym$ extracts the symmetric part, $\Sym(\mat{Z}) = \frac{\mat{Z} + \mat{Z}^T}{2}$ where $\mat{Z}$ is a square matrix.

\section{Algorithms}
\subsection*{Gradient descent schemes}
In the previous section, we have developed all the necessary ingredients to propose gradient descent schemes.  If $(\mat{G}, \mat{H})$ is the current iterate, we take a simultaneous step in both the variables in the negative Riemannian gradient direction with the new metric (\ref{eq:metric_gh}) $(\mat{G}, \mat{H})$, i.e., $(\mat{SH}(\mat{H}^T\mat{H})^{-1},\quad \mat{S}^T \mat{G}(\mat{G}^T\mat{G})^{-1}  )$ where $\mat{S} = \frac{2}{|\Omega|} \left ( \mathcal{P}_{\Omega}(\mat{GH}^T) - \mathcal{P}_{\Omega}(\widetilde{\mat{X}}) \right ) $ is the Euclidean gradient in the space $\mathbb{R}^{n \times m}$. Note that $\frac{\partial}{\partial \mat{G}}  = \mat{SH}$ (obtained by chain rule) and the scaled version is post multiplied by $(\mat{H}^T\mat{H})^{-1}$. Similarly, $\frac{\partial}{\partial \mat{H}}  = \mat{S}^T\mat{G}$ and the scaled version is post multiplied by $(\mat{G}^T\mat{G})^{-1}$. The update rule, thus, is
\begin{equation}\label{eq:gradient_descent}
\begin{array}{llll}
\mat{G}_+ & = & \mat{G} - t\mat{SH}(\mat{H}^T\mat{H})^{-1} \\
\mat{H}_+ & = & \mat{H} - t\mat{S}^T \mat{G}(\mat{G}^T\mat{G})^{-1} \\
\end{array}
\end{equation}
with the step-size $t>0$. In our implementation, we perform an Armijo linesearch with an adaptive step-size procedure \cite{mishra12a} to update $t$ .

Similarly, a geometric conjugate gradient (CG) can also be described.  We implement Algorithm $13$ of \cite{absil08a} with the tools developed here.  Once a conjugate gradient direction is obtained, the optimal step-size is then computed by solving (\ref{eq:exact_linesearch}).

\subsection*{Trust-region scheme}
In a trust-region scheme we first build a locally second-order model of the objective function in a neighborhood and minimize the model function to obtain a candidate iterated. This is called the trust-region subproblem. Depending on the obtained and the desired decrease, the neighborhood is modified \cite{nocedal06a}. 

Analogous to trust-region algorithms in the Euclidean space, trust-region algorithms on a quotient manifold with guaranteed quadratic rate convergence have been proposed in \cite{absil07a, absil08a}. In particular, the trust-region subproblem on the abstract space $\mathcal{M}_r$ is horizontally lifted to $\mathcal{H}_{\bar{x}} \overline{\mathcal{M}}_r$ and	formulated as 
\begin{equation}\label{eq:trust_region_subproblem}
\begin{array}{lll}
\min_{{\bar \eta} \in \mathcal{H}_{\bar{x}} \overline{\mathcal{M}}_r  } & \quad {\bar \phi} ({\bar x}) +  g_{\bar x} (  {\bar \eta} ,   \overline{\grad \phi(x)} ) + \frac{1}{2}  g_{x} ( {\bar \eta}, \overline{\hess \phi(x) [\eta] }   ) \\
\subject & {g}_{\bar x}  ( {\bar \eta} , {\bar \eta}) \leq \delta.
\end{array}
\end{equation}
where  $\delta$ is the trust-region radius and $\overline{\grad\phi}$ and $\overline{\hess \phi}$ are the horizontal lifts of the Riemannian gradient and its directional derivative (covariant derivative in the direction $\eta$ denoted by $\hess \phi  [\eta] $) of the least square objective function $\phi$ on $\mathcal{M}_r$. Here $\bar{\phi}_{\bar x} = \frac{ 1}{|\Omega|}\|\mathcal{P}_{\Omega}(\mat{GH}^T) - \mathcal{P}_{\Omega}(\widetilde{\mat{X}})\|_F^2$ at $\bar{x} = (\mat{G}, \mat{H})$. 

The covariant derivative $\hess \phi  [\eta] $ is computed using the relations (\ref{eq:Riemannian_connection}) and (\ref{eq:connection_total_space})
\[
\overline{\hess \phi  [\eta]}  = \overline{\nabla_{\eta} \grad \phi } = \Pi_{\bar x} \left( \overline{\nabla}_{\overline{\eta}} \overline{\grad \phi} \right ).
\]
The trust-region subproblem is solved efficiently using the generic solver GenRTR \cite{genrtr}. Refer \cite{absil07a, genrtr} for the algorithmic details. The output is a direction $\bar{\eta}$ that minimizes the model. We initialize the trust-region radius $\delta$ 
\begin{equation}\label{eq:delta}
\delta_0  = t_0 \| \overline{\grad\phi(x_0) }  \|_ {\overline{x_0}}  \\
\end{equation} 
where $t_0 >0$ is the optimal stepsize in the negative $\grad \phi(x_0) $ direction at the initial iterate $\overline{x_0} \in \overline{\mathcal{M}}_r$. The exact line-search procedure has been described earlier. The overall bound on the trust-region radii, $\bar{\delta}$ is fixed at $2^{10}\delta_0$ (relatively a large number, we allow $10$ updates). For subsequent iterates $\delta$ is updated as in Algorithm $4.1$ in \cite{nocedal06a}.

\subsection*{Connection with LMaFit}
The LMaFit algorithm \cite{wen10a} for the low-rank matrix completion problem has been a popular benchmark owing to simpler updates of iterates and tuned step-size updates in turn leading to a superior time per iteration complexity. The algorithm relies on the factorization $\mat{X} = \mat{GH}^T$ to alternatively learn the matrices $\mat{X}$, $\mat{G}$ and $\mat{H}$ so that the error $\|\widetilde{ \mat{X}}  - \mat{GH}^T \|^2_F$ is minimized while ensuring that the entries of $\mat{X}$ agree with the known entries i.e., $\mathcal{P}_{\Omega}(\mat{X}) = \mathcal{P}_{\Omega}(\widetilde{\mat{X}})$. The algorithm is a tuned version of the block-coordinate descent algorithm in the product space that has a superior computational cost per iteration and better convergence than the straight forward non-linear Gauss-Seidel scheme (GS algorithm). We show here both the non-linear GS and LMaFit updates.

The GS algorithm is an alternating minimization scheme, when each variable is updated first and used in the update of the other variables. However, instead of sequential updates of the variables, a simultaneous update of the variables has the following form. Given an iterate $(\mat{G}, \mat{H})$, simultaneously updating the variables to $(\mat{G}_+, \mat{H}_+)$
\begin{equation}\label{eq:GS_updates}
\begin{array}{lll}
\mat{G}_+ & = & \mat{G} - \mat{SH}(\mat{H}^T\mat{H})^{-1} \\
\mat{H}_+ & = & \mat{H} - \mat{S}^T \mat{G}(\mat{G}^T\mat{G})^{-1} \\
\end{array}
\end{equation}
where $\mat{S} = \frac{2}{|\Omega|} \left ( \mathcal{P}_{\Omega}(\mat{GH}^T) - \mathcal{P}_{\Omega}(\widetilde{\mat{X}}) \right )$ is the Euclidean gradient of the objective function (\ref{eq:matrix-completion-formulation}) in the space $\mathbb{R}^{n \times m}$. As mentioned before, LMaFit is a tuned version of the GS updates (\ref{eq:GS_updates}), the updates are written as 
\begin{equation}\label{eq:lmafit}
\begin{array}{llll}
\mat{G}_+ = (1- \omega)\mat{G} + \omega( \mat{G} - \mat{SH}(\mat{H}^T\mat{H})^{-1}   ) \\
\mat{H}_+ = (1- \omega)\mat{H} + \omega( \mat{H} - \mat{S}^T \mat{G}(\mat{G}^T\mat{G})^{-1}   )
\end{array}
\end{equation} 
where the weight $\omega > 1$ is updated in an efficient way \cite{wen10a}. The connection between (simultaneous version of) LMaFit and our gradient descent algorithm is straight forward by looking at the similarity between (\ref{eq:gradient_descent}) and (\ref{eq:lmafit}). LMaFit is a particular case of our geometric gradient descent algorithm (\ref{eq:gradient_descent}).

\section{Numerical simulations}
We compare our algorithms with those from the right-invariant geometry \cite{mishra12a}, LMaFit \cite{wen10a} and LRGeom \cite{vandereycken11a}. Unlike the right-invariant geometry and LMaFit, LRGeom is based on the \emph{embedded} geometry which is different from the geometries based on a factorization model. This view simplifies some notions of  the geometric objects that can be interpreted in a straight forward way. In the matrix completion case, this allows to compute the initial guess for the Armijo line-search efficiently \cite{vandereycken11a}.

For the gradient descent implementations of our geometry and the right-invariant geometry we use the Armijo lineseach procedure  with the adaptive step-size update proposed in \cite{mishra12a}. For the gradient descent implementation of LRGeom, we use their step-size guess at each iteration. 

For the CG implementations, we use the robust ${\rm P-R}_+$ update for the parameter $\beta$ \cite{nocedal06a}. We use an exact lineseach procedure for both our new geometry and the right-invariant geometry. For the CG implementation of LRGeom, we use their step-size guess for each iteration.

For trust-region algorithms, we also compare with algorithms based on our new geometry, the right-invariant geometry in \cite{mishra12a} and the embedded geometry in \cite{vandereycken11a}. The trust-region subproblem is solved using the GenRTR package \cite{genrtr} and the trust-region radius for the algorithms based on the new geometry and the right-invariant geometry are initialized as in (\ref{eq:delta}). 


All simulations are performed in MATLAB on a $2.53$ GHz Intel Core $\rm{i}5$ machine with $4$ GB of RAM. We use MATLAB codes of the competing algorithms for our numerical studies. The codes are available from the respective authors' webpages. For each example, a $n \times m$ random matrix of rank $r$ is generated according to a Gaussian distribution with zero mean and unit standard deviation and a fraction of the entries are randomly removed with uniform probability. The dimensions of $n \times m$ matrices of rank $r$ is $(n + m - r)r$. The over-sampling (OS) ratio determines the number of entries that are known. A $\rm{OS} = 6$ means that $6(n + m - r)r$ number of randomly and uniformly selected entries are known a priori out of $nm$ entries. Numerical codes for the proposed algorithms for the new geometry are available from the first author's homepage.

Three instances are considered where we demonstrate the efficacy of our framework. We randomly initialize all the competing algorithms. The algorithms are terminated once the cost, $\frac{ 1}{|\Omega|}\|\mathcal{P}_{\Omega}(\mat{X}) - \mathcal{P}_{\Omega}(\widetilde{\mat{X}})\|_F^2 $ is below $10^{-20}$ or the number of iterations exceeds $500$.

\subsection{Comparison with gradient schemes} \label{sec:comparison_gd}

\subsubsection*{A large-rank instance}
\begin{figure}[htp]
\subfigure{
\includegraphics[scale = .32]{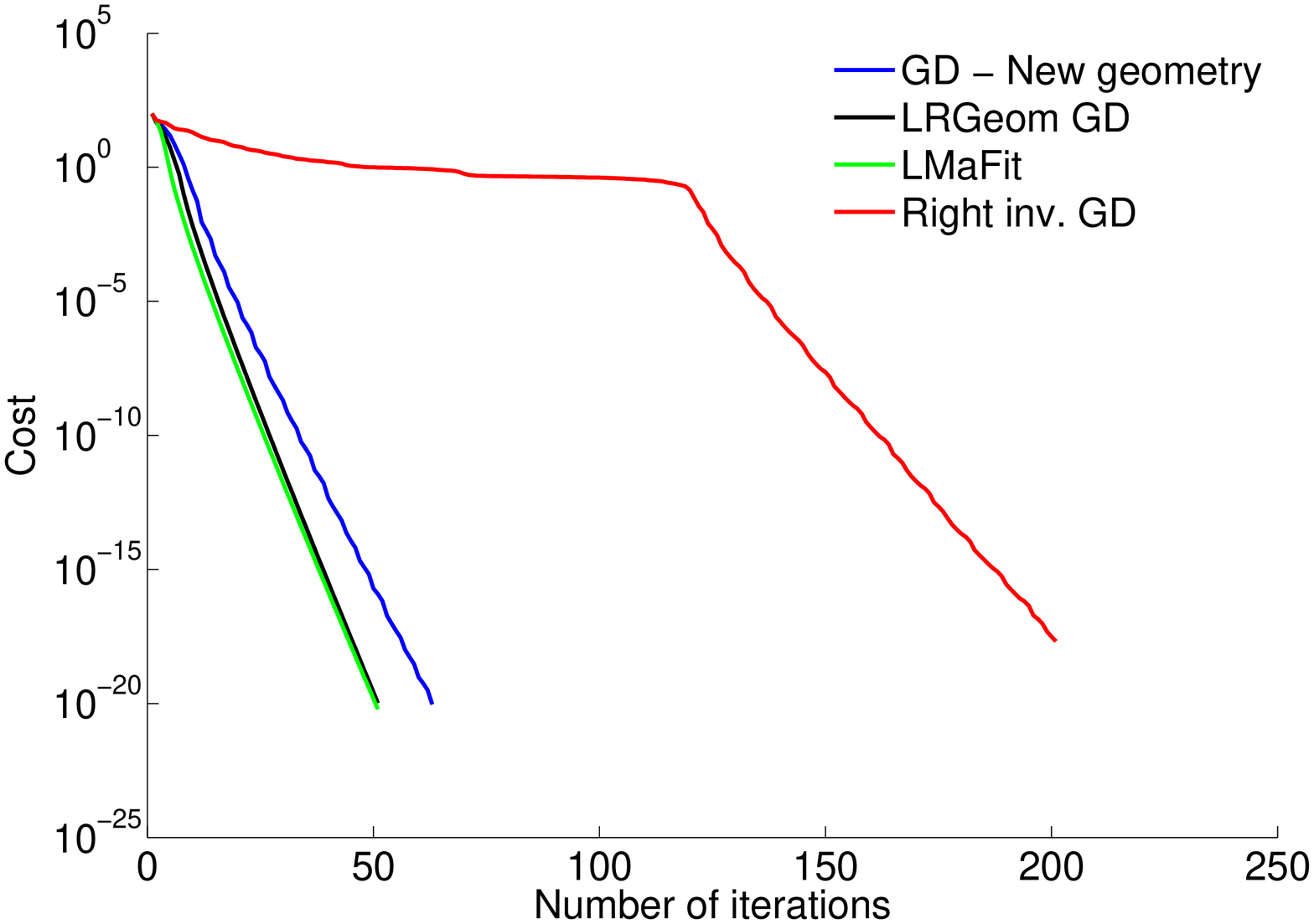}
}
\subfigure{
\includegraphics[scale = .32]{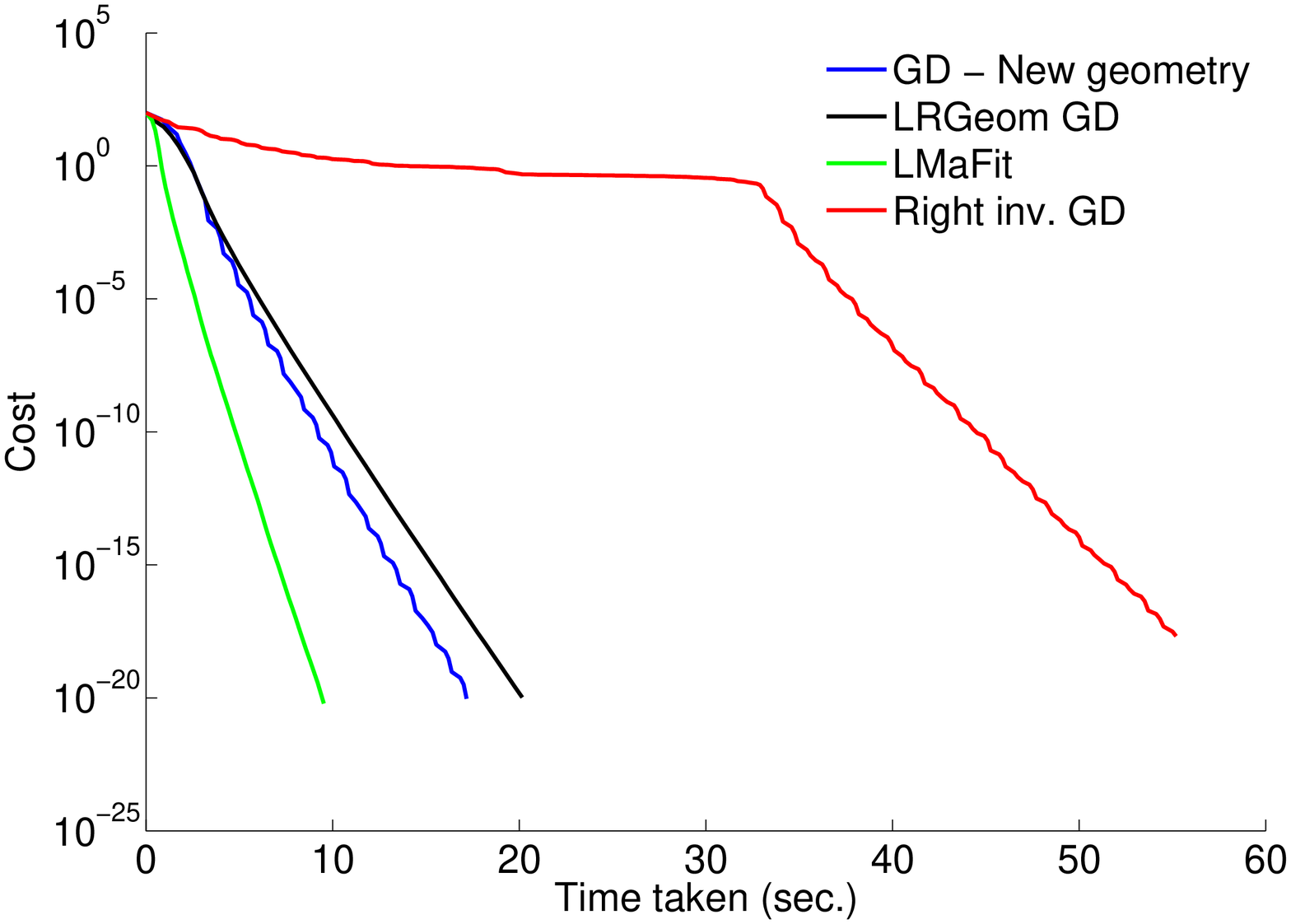}
}
\subfigure{
\includegraphics[scale = .32]{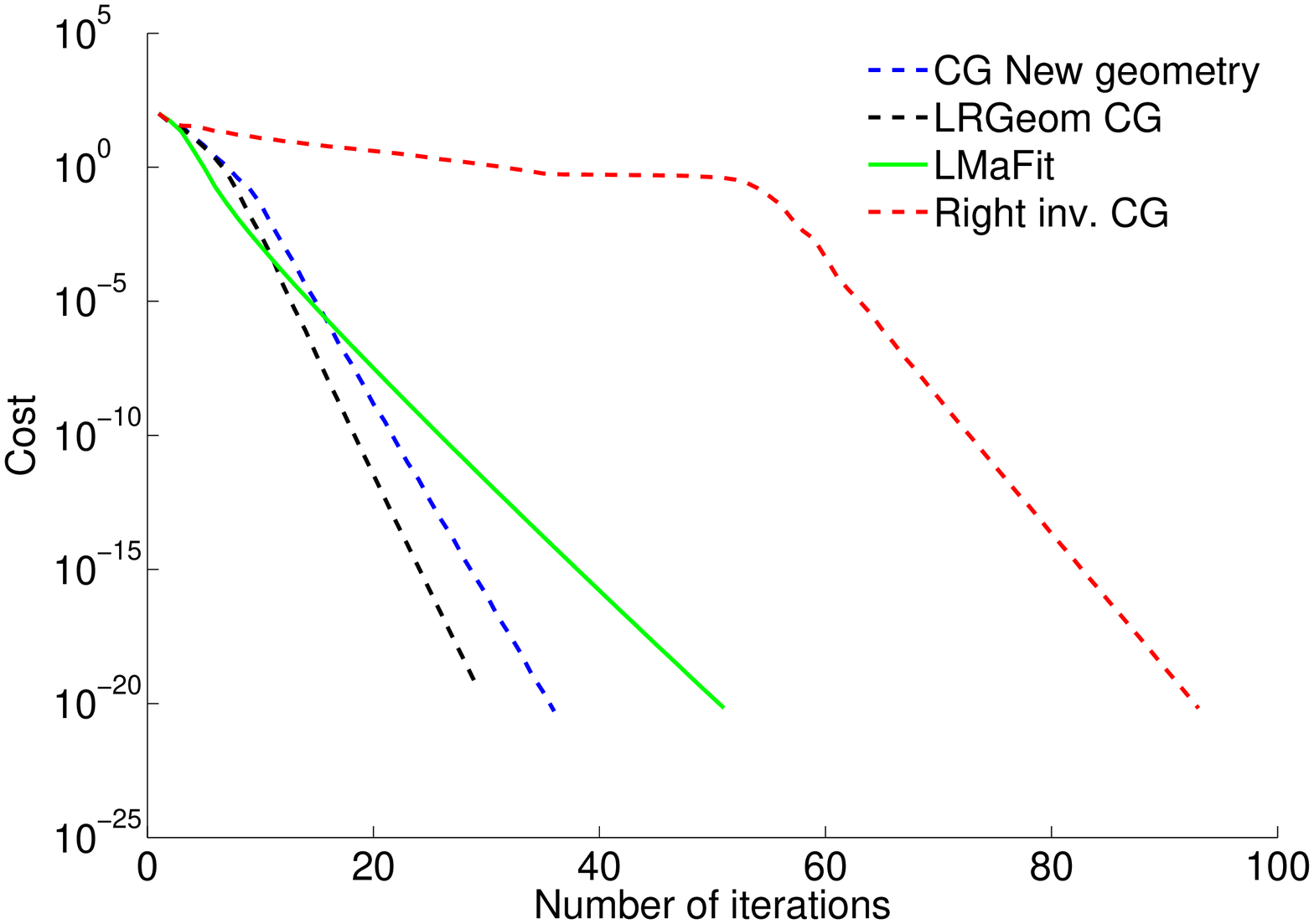}
}
\subfigure{
\includegraphics[scale = .32]{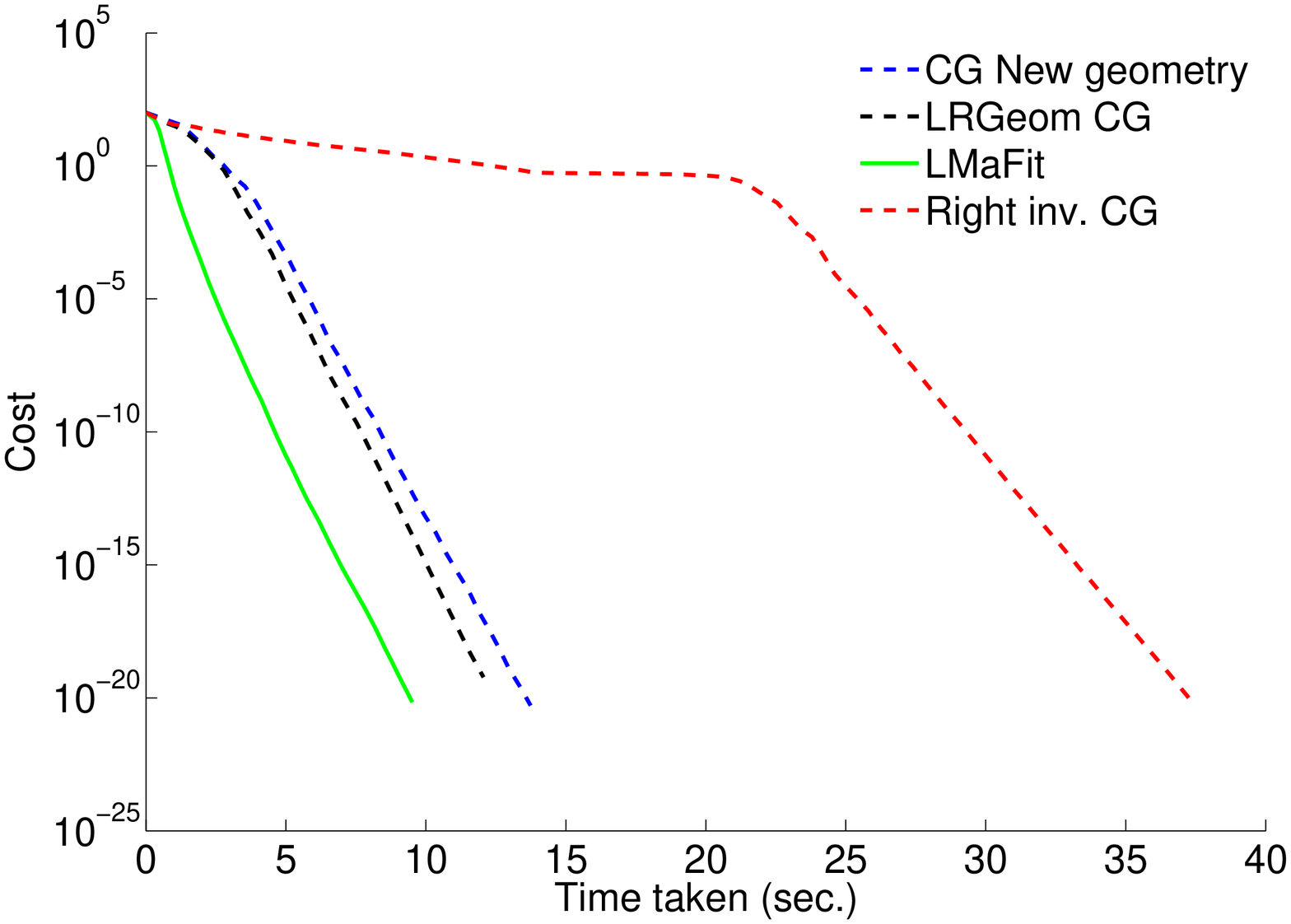}
}
\caption{LMaFit performs extremely well when a large number of entries are known. The instance considered here involves $n = m = 1000$ and $r = 50$ completion with ${\rm OS = 5}$. The step-size parameter of LMaFit is properly tuned for these instances. \emph{Top:} Gradient descent algorithms. \emph{Bottom: } Conjugate gradient algorithms. The right-invariant geometry performs poorly with random initialization.}
\label{fig:large_rank_instance}
\end{figure}
Consider a case when $n = m = 1000$ and $r = 50$ with ${\rm OS = 5}$, which implies that we know $49\%$ of the entries (which is a big number). In this case, one would expect LMaFit to perform better than both gradient descent and CG algorithms primarily because the parameter $\omega$ is properly tuned. As more and more entries are known, the completion problem (\ref{eq:matrix-completion-formulation}) approaches the matrix approximation problem (\ref{eq:matrix_approximation}) problem. One would also expect the right-invariant geometry of \cite{meyer11a} to perform poorly on this instance with random initialization as it neglects scaling that is required (\ref{eq:difference_metric}) \footnote{With an initialization proposed in \cite{keshavan10a}, the poor performance of the right-invariant algorithms diminishes.}. In fact, the right-invariant geometry \emph{inversely} scales the gradient (\ref{eq:difference_metric}). The plots are shown in Figure \ref{fig:large_rank_instance}.

\subsubsection*{Small-rank and still-smaller-rank instances}
\begin{figure}[H]
\subfigure{
\includegraphics[scale = .32]{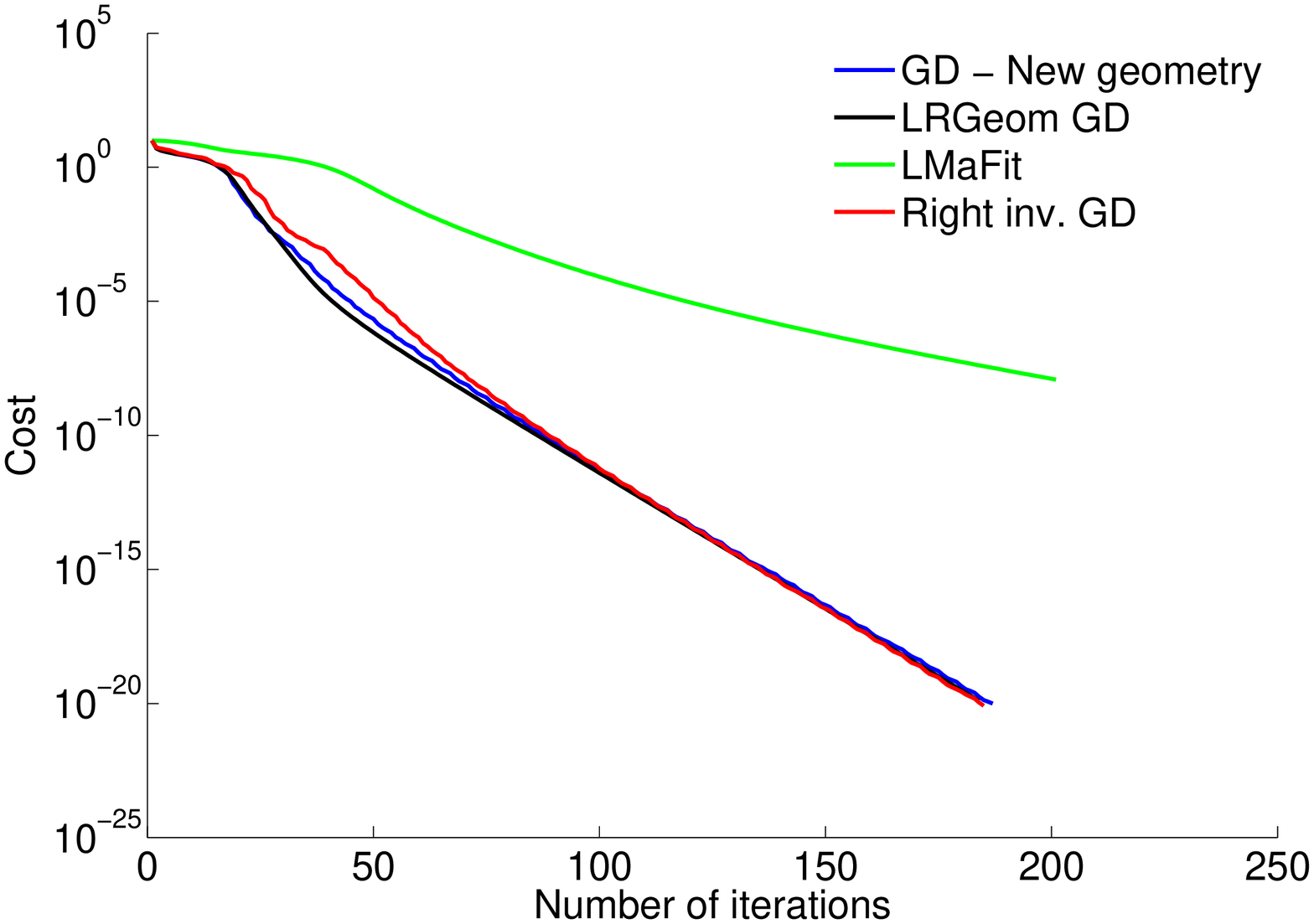}
}
\subfigure{
\includegraphics[scale = .32]{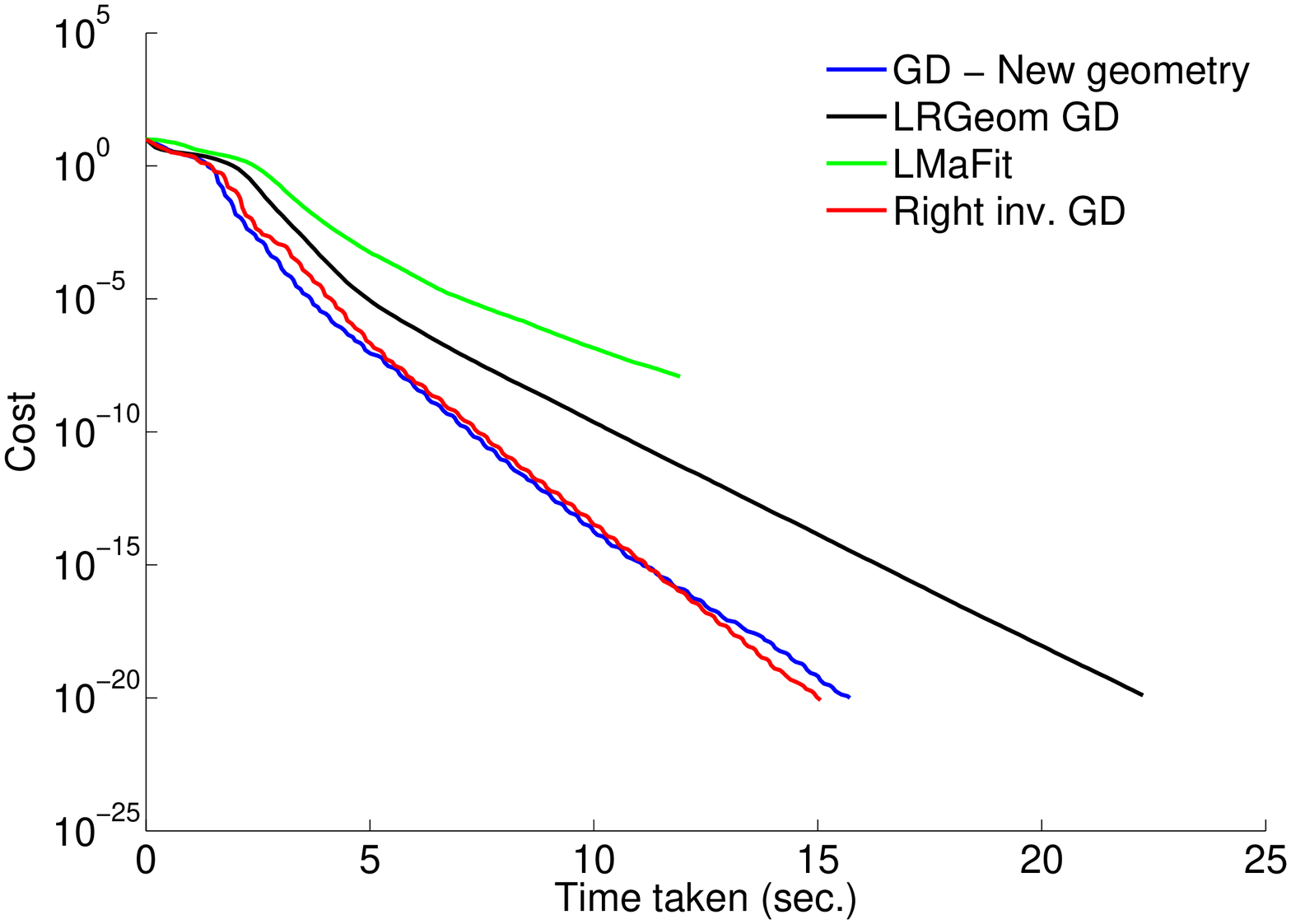}
}
\subfigure{
\includegraphics[scale = .32]{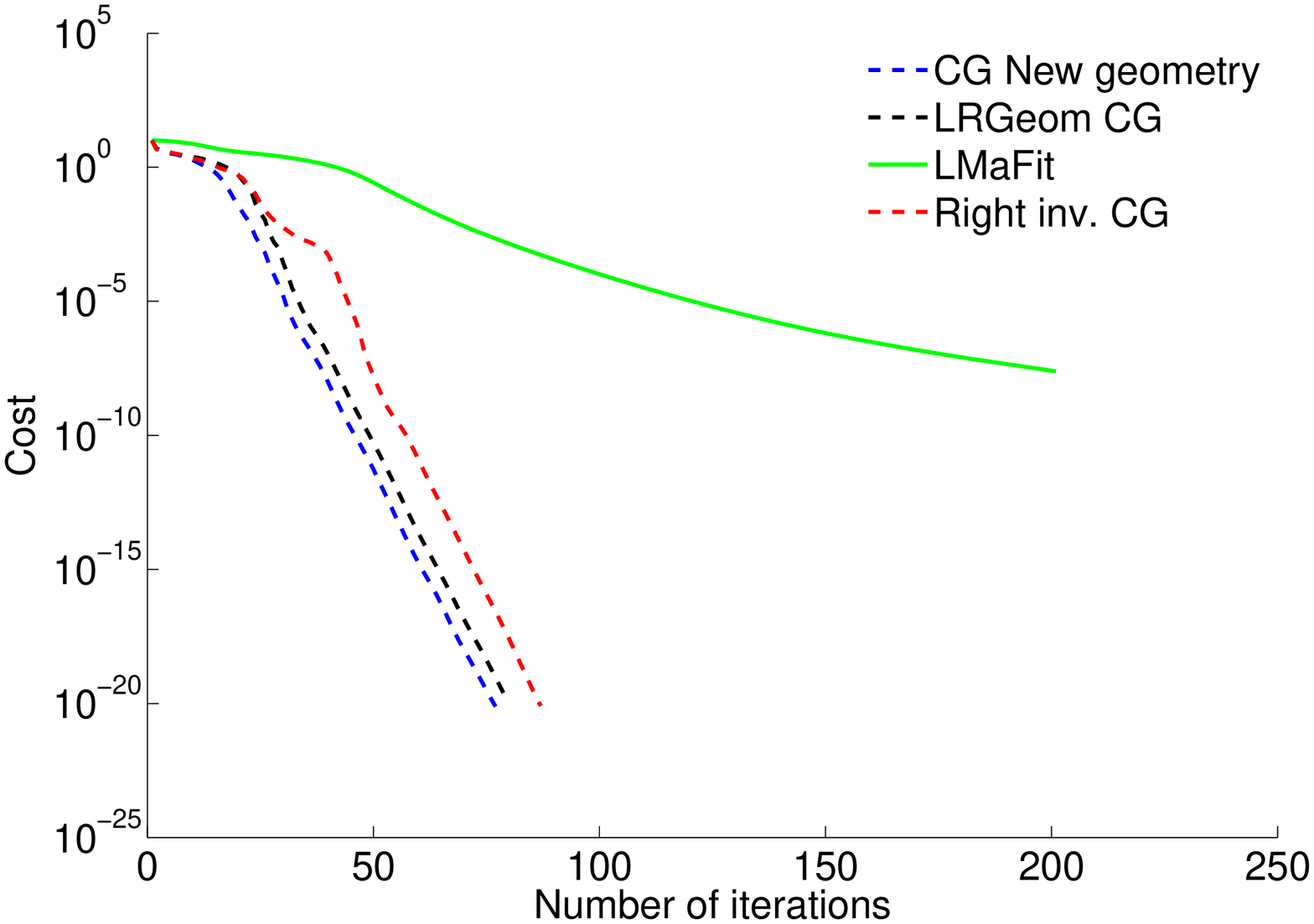}
}
\subfigure{
\includegraphics[scale = .32]{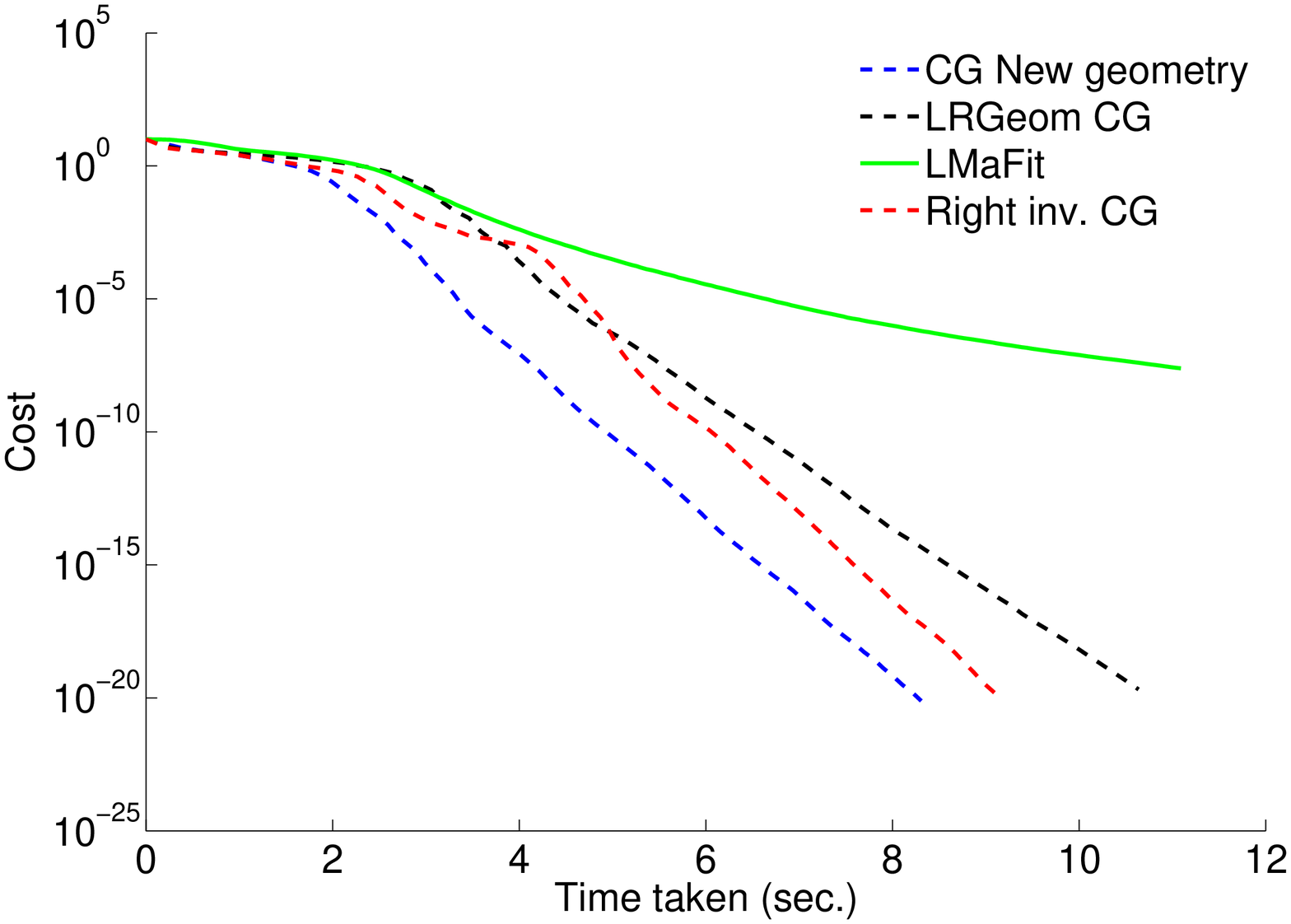}
}
\caption{Competitive performance our geometry both for CG and gradient descent schemes. $n = m =10000$ and $r = 5$ with ${\rm OS = 5}$. LMaFit performs poorly for these instances as the parameter tuning is not proper. \emph{Top:} Gradient descent algorithms. \emph{Bottom: } Conjugate gradient algorithms.}
\label{fig:small_rank_instance}
\end{figure}
Consider the case $n = m = 10000$ and $r = 5$ with ${\rm OS = 5}$, which implies that we know $0.5\%$ of the entries (which is a small number). The plots are shown in Figure \ref{fig:small_rank_instance}. LMaFit in this case suffers from poor convergence because of the poor tuning of the parameter $\omega$. All the gradient descent algorithms have a similar rate of convergence however, timing-wise the gradient descent implementations of our new geometry and right-invariant geometry have better timing than the gradient descent implementation of LRGeom, suggesting that the adaptive linesearch is competitive in a gradient descent setting. CG algorithms based on our new geometry outperform others.
 
\begin{figure}[H]
\subfigure{
\includegraphics[scale = .32]{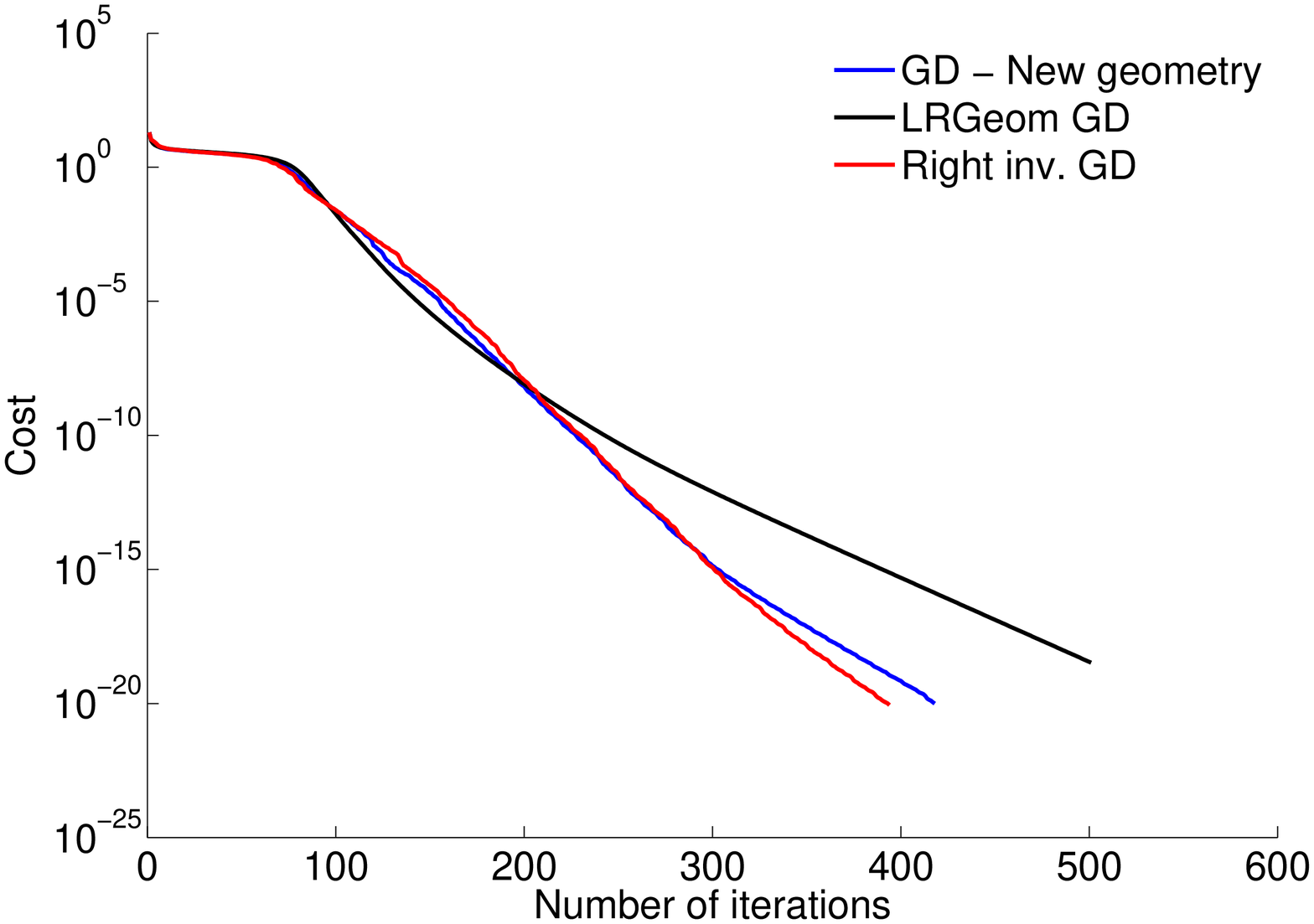}
}
\subfigure{
\includegraphics[scale = .32]{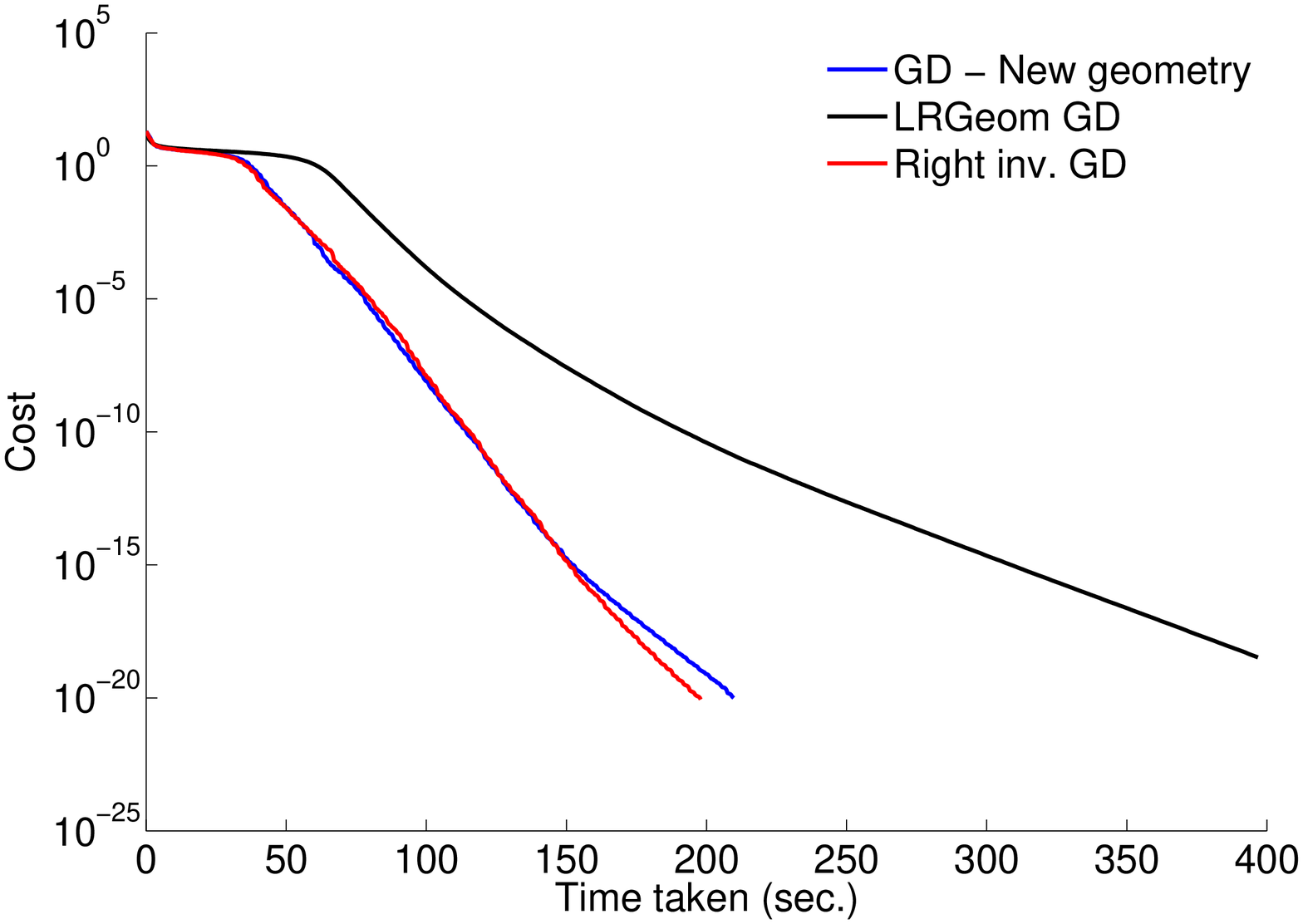}
}
\subfigure{
\includegraphics[scale = .32]{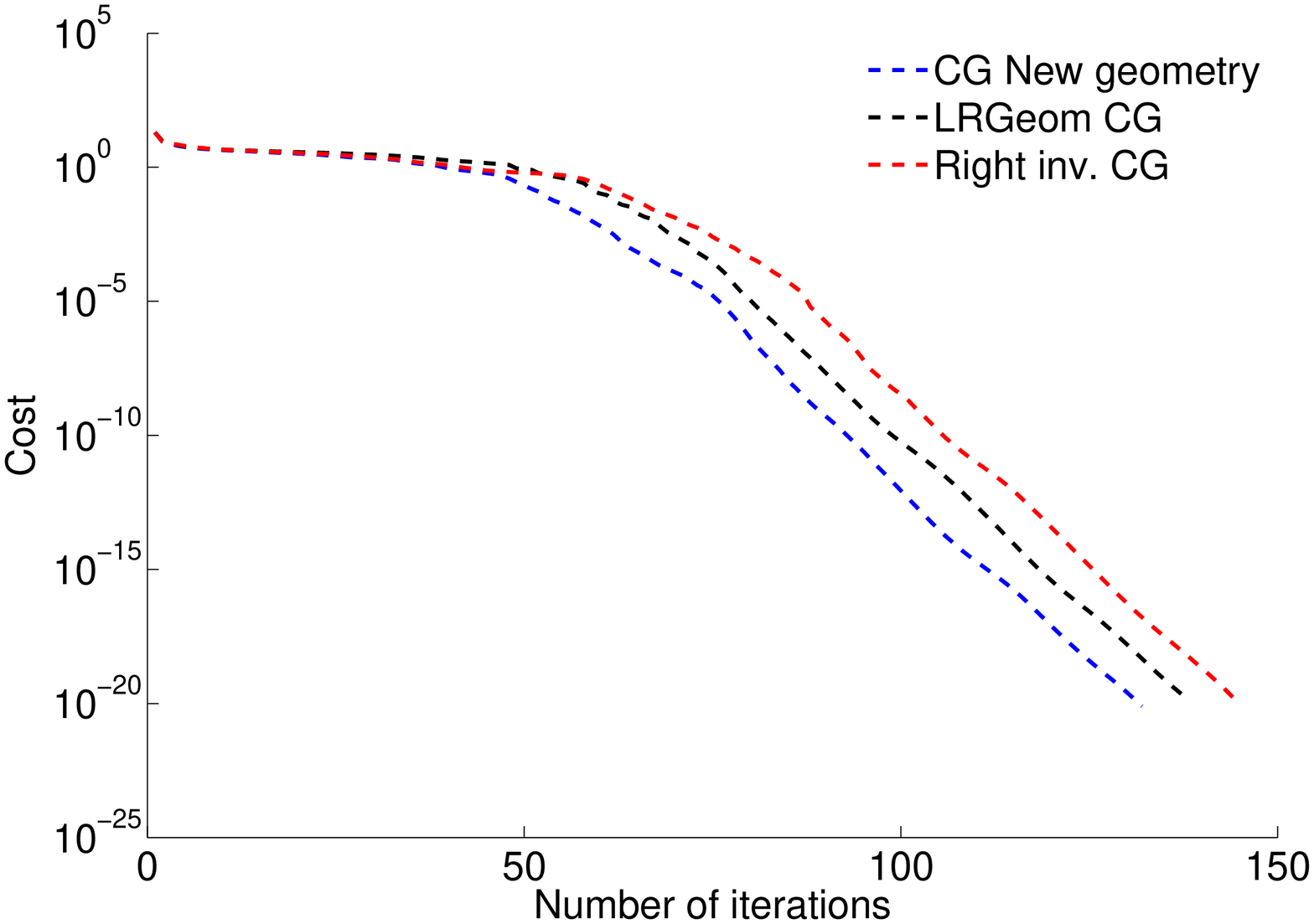}
}
\subfigure{
\includegraphics[scale = .32]{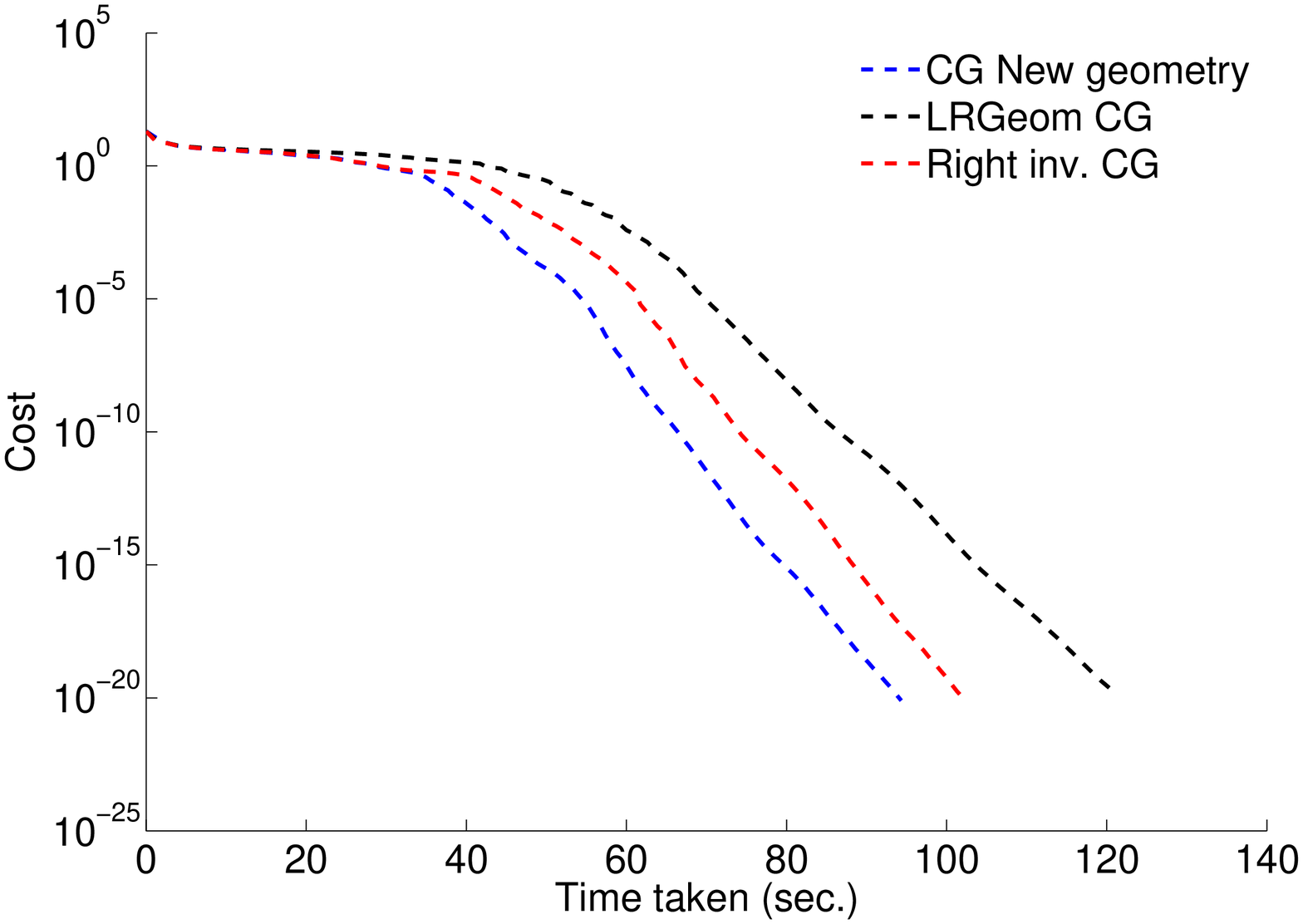}
}
\caption{Competitive performance our geometry on a problem instance, $n = m = 32000$ and $r = 10$ with ${\rm OS = 3}$. Our CG scheme outperforms others. \emph{Top:} Gradient descent algorithms. \emph{Bottom: } Conjugate gradient algorithms.}
\label{fig:smaller_rank_instance}
\end{figure}

Consider a still bigger example of $n = m = 32000$ and $r = 10$ with ${\rm OS = 3}$. In this case we know $0.12\%$ of the entries (which is a very small number). The performance of different algorithms are shown in Figure \ref{fig:smaller_rank_instance}. Our CG scheme convincingly outperforms other CG schemes. The gradient descent schemes our geometry and the right-invariant geometry seem to perform similarly.

\subsection{Comparison with trust-region algorithms}
We consider a large scale instance of size $32000 \times 32000$ of rank $10$. The number of entries are uniformly revealed with ${\rm OS} = 5$. All the algorithms are initialized similarly as in \cite{keshavan10a}, by taking the dominant SVD of the sparse incomplete matrix\footnote {The TR algorithms do not seem to compete favorably with the conjugate gradient algorithms for a number of instances when initialized randomly.}.

\begin{figure}[H]
\subfigure{
\includegraphics[scale = .32]{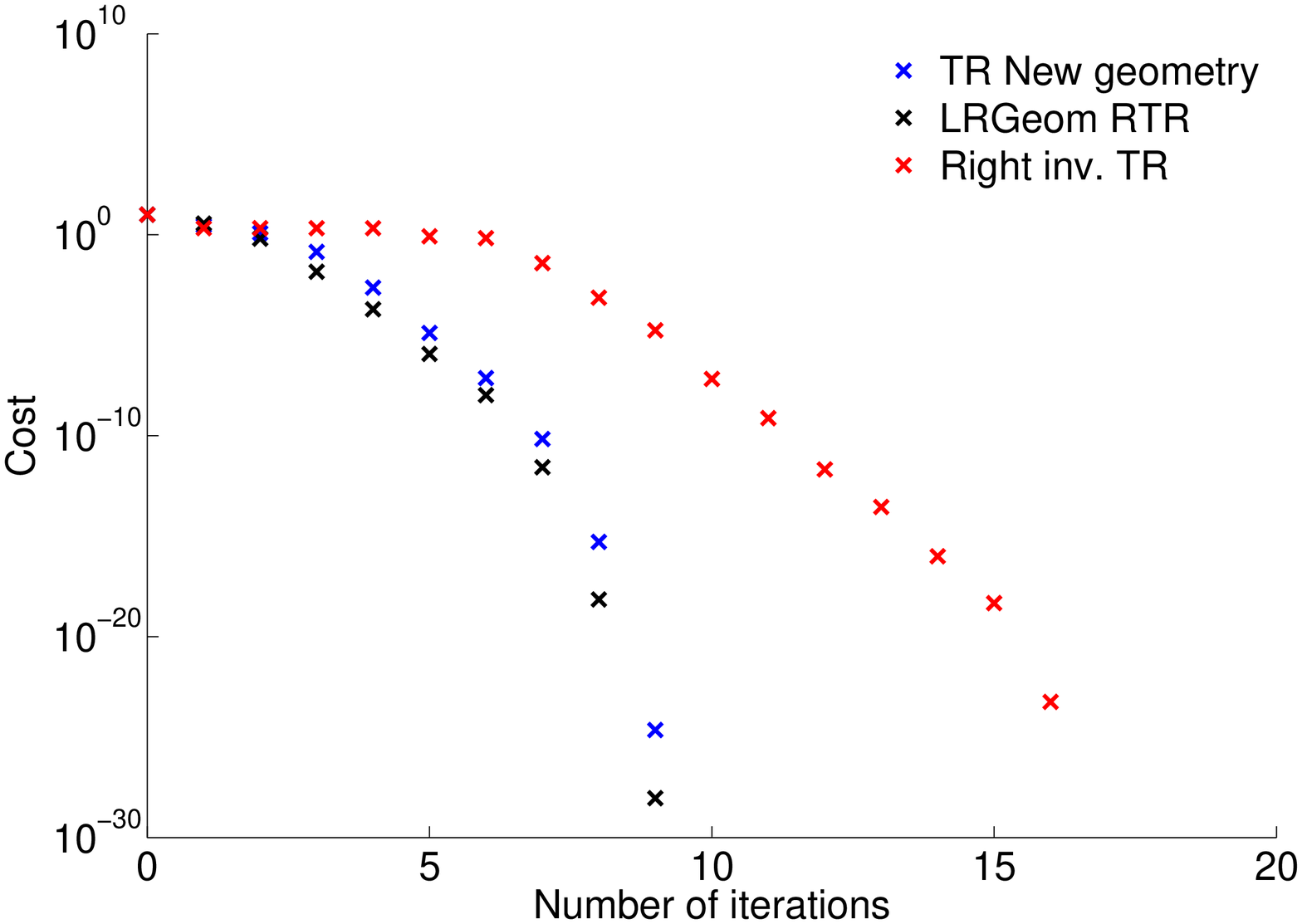}
}
\subfigure{
\includegraphics[scale = .32]{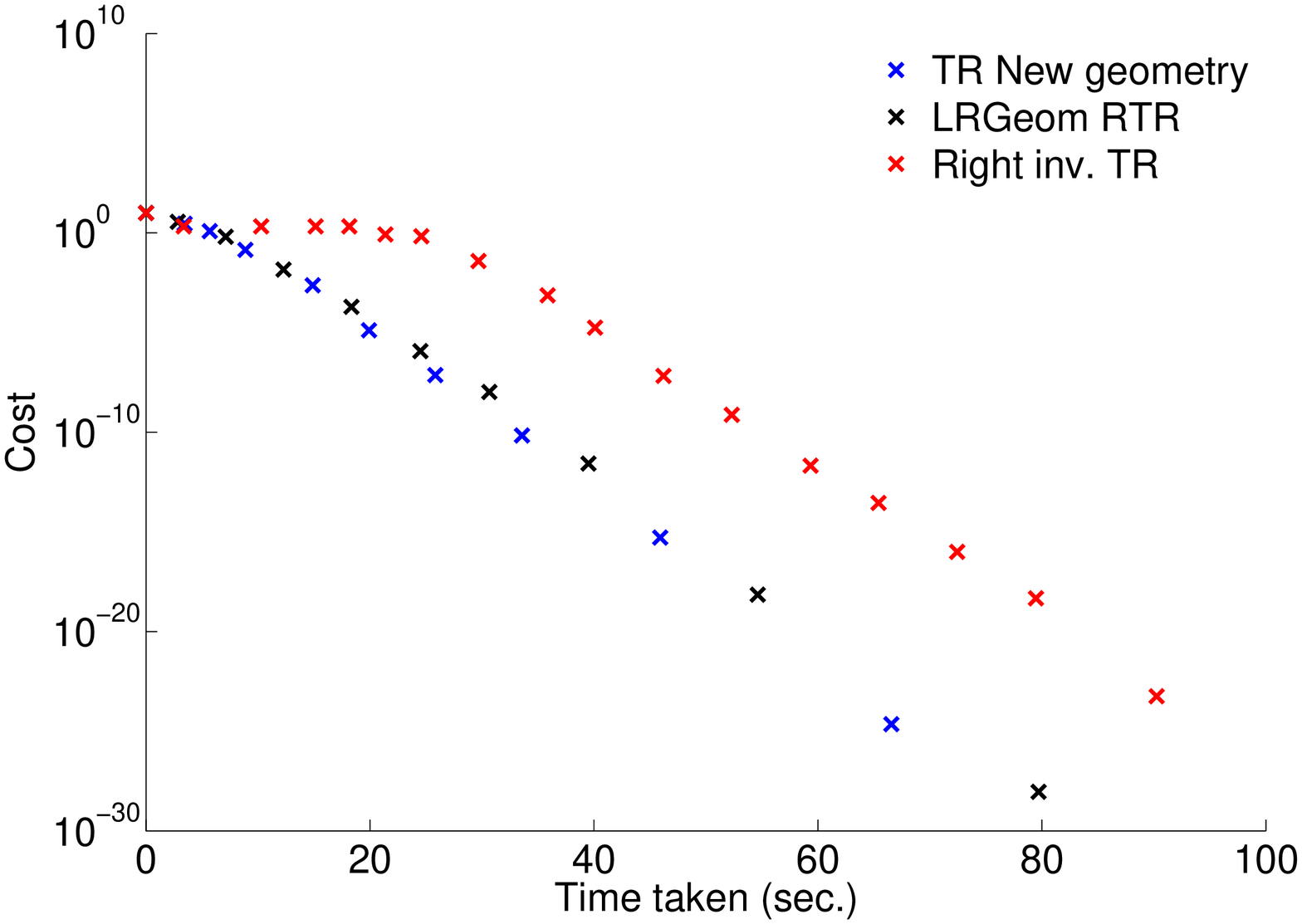}
}
\caption{$n = m =32000$ and $r = 10$ with ${\rm OS = 5}$. Our trust-region scheme is clearly superior to that of the right-invariant geometry. Also our trust-region algorithm competes favorably with that of the embedded geometry in a number of situations.}
\label{fig:TR}
\end{figure}

The plots in Figure \ref{fig:TR} show a favorable performance of our trust-region scheme. As expected, our TR algorithm is superior to that of the right-invariant geometry.  With respect to the embedded approach, the performance of the both the algorithms are similar both in the number of outer (as well as inner iterations for the trust-region subproblem) and in computational burden per iteration.

\section{Conclusion}
We have described a new geometry for the set of fixed-rank matrices. The geometry builds on the framework of quotient manifold but the new metric (\ref{eq:metric_gh}) additionally exploits the particular cost function of the low-rank matrix completion problem. The proposed algorithms connect to the state-of-the-art LMaFit algorithm, which is viewed as a tuned (step-size) algorithm in our framework. We have developed all the necessary ingredients to perform first-order and second-order optimization with the new geometry. We have shown that an exact linesearch is numerically feasible with our choice of updating low-rank factors. The resulting class of algorithms compete favorably with the state-of-the-art in a number of problem instances.

\bibliographystyle{amsalpha}
\bibliography{GS_gh}



\end{document}